\documentclass[10pt,twocolumn,letterpaper]{article}

\usepackage[accsupp]{axessibility} 
\usepackage{iccv}
\usepackage{times}
\usepackage{epsfig}
\usepackage{graphicx}
\usepackage{amsmath}
\usepackage{amssymb}
\usepackage{bm}
\usepackage{diagbox}
\usepackage{multirow}
\usepackage{booktabs}
\usepackage{bbding}
\usepackage{subfigure}
\usepackage{soul}
\usepackage{xcolor}

\makeatletter
\newcommand{\rmnum}[1]{\romannumeral #1}
\newcommand{\Rmnum}[1]{\expandafter\@slowromancap\romannumeral #1@}
\makeatother

\usepackage[pagebackref=true,breaklinks=true,letterpaper=true,colorlinks,bookmarks=false,citecolor=purple]{hyperref}

\iccvfinalcopy % *** Uncomment the

\def\iccvPaperID{1418} % *** Enter the ICCV Paper ID here

\ificcvfinal\fi

\begin{document}

%%%%%%%%% TITLE
\title{Spatio-Temporal Domain Awareness for Multi-Agent Collaborative Perception}

\author{Kun Yang$^{1}$\footnotemark[2] $\quad$
        Dingkang Yang$^{1}$\footnotemark[2] $\quad$
        Jingyu Zhang$^{1}$  $\quad$
        Mingcheng Li$^{1}$  $\quad$
        Yang Liu$^{1}$  $\quad$ \\
        Jing Liu$^{1}$  $\quad$ 
        Hanqi Wang$^{1}$  $\quad$ 
        Peng Sun$^{2}$\footnotemark[1]  $\quad$
        Liang Song$^{1}$\footnotemark[1] $\quad$ \\ 
        $^{1}$Academy for Engineering and Technology, Fudan University$\quad$
        $^{2}$Duke Kunshan University\\
{\tt\small $\{$kunyang20,dkyang20,songl$\}$@fudan.edu.cn}
}
\maketitle
\ificcvfinal\thispagestyle{empty}\fi

\renewcommand{\thefootnote}{\fnsymbol{footnote}} 
\footnotetext[2]{Equal contributions in no particular order. The work was done when Dingkang Yang was at the Institute of Meta-Medical, IPASS. The two first authors thank Runsheng Xu for providing constructive suggestions.} 
\footnotetext[1]{Corresponding authors.} 

%%%%%%%%% ABSTRACT
\begin{abstract}

   Multi-agent collaborative perception as a potential application for vehicle-to-everything communication could significantly improve the perception performance of autonomous vehicles over single-agent perception. However, several challenges remain in achieving pragmatic information sharing in this emerging research. In this paper, we propose {\tt SCOPE}, a novel collaborative perception framework that aggregates the spatio-temporal awareness characteristics across on-road agents in an end-to-end manner. Specifically, {\tt SCOPE} has three distinct strengths: i) it considers effective semantic cues of the temporal context to enhance current representations of the target agent; ii) it aggregates perceptually critical spatial information from heterogeneous agents and overcomes localization errors via multi-scale feature interactions; iii) it integrates multi-source representations of the target agent based on their complementary contributions by an adaptive fusion paradigm. To thoroughly evaluate {\tt SCOPE}, we consider both real-world and simulated scenarios of collaborative 3D object detection tasks on three datasets. Extensive experiments show the superiority of our approach and the necessity of the proposed components. The project link is \url{https://ydk122024.github.io/SCOPE/}.
\end{abstract}

%%%%%%%%% BODY TEXT
\section{Introduction}
\label{intro}

Perceiving complex driving environments is essential for ensuring road safety~\cite{ram2016effect} and traffic efficiency~\cite{luo2022complementarity} in autonomous driving. Despite the rapid development in the perception ability of intelligent vehicles through deep learning technologies~\cite{du2021learning,el2019rgb,lang2019pointpillars,yang2022emotion,yang2023target}, the conventional single-agent perception paradigm remains challenging. Single-agent perception system is generally vulnerable and defective due to the inevitable practical difficulties on the road, such as restricted detection range and occlusion~\cite{yuan2021temporal,zhang2021safe}. To alleviate the inadequate observation constraints of isolated agents, multi-agent collaborative perception has emerged as a promising solution in recent years~\cite{li2021learning,xiang2023v2xp}. With Vehicle-to-Everything (V2X) oriented communication applications, collaborative perception effectively promotes perspective complementation and information sharing among agents.
% resulting in a more precise perception of driving scenarios.

Based on recent LiDAR-based 3D collaborative perception datasets~\cite{xu2023v2v4real,xu2022v2x,xu2022opv2v,yu2022dair}, several impressive works have achieved a trade-off between collaborative perception performance and communication bandwidth of Autonomous Vehicles (AVs) via well-designed mechanisms. These remarkable efforts include knowledge distillation-based intermediate collaboration~\cite{li2021learning}, handshake communication mechanism~\cite{liu2020when2com}, V2X-visual transformer~\cite{xu2022v2x}, and spatial information selection strategy~\cite{hu2022where2comm}. However, there is still considerable room for progress in achieving a pragmatic and effective collaborative perception. 
Concretely, existing methods invariably follow a single-frame static perception pattern, suffering from the data sparsity shortcoming of 3D point clouds~\cite{yuan2021temporal} and ignoring meaningful semantic cues in the temporal context~\cite{zhou2022centerformer}. The insufficient representation of the moving objects in the current frame potentially restricts the perception performance of target vehicles.

 \begin{figure*}[t]
  \centering
  \includegraphics[width=0.9\linewidth]{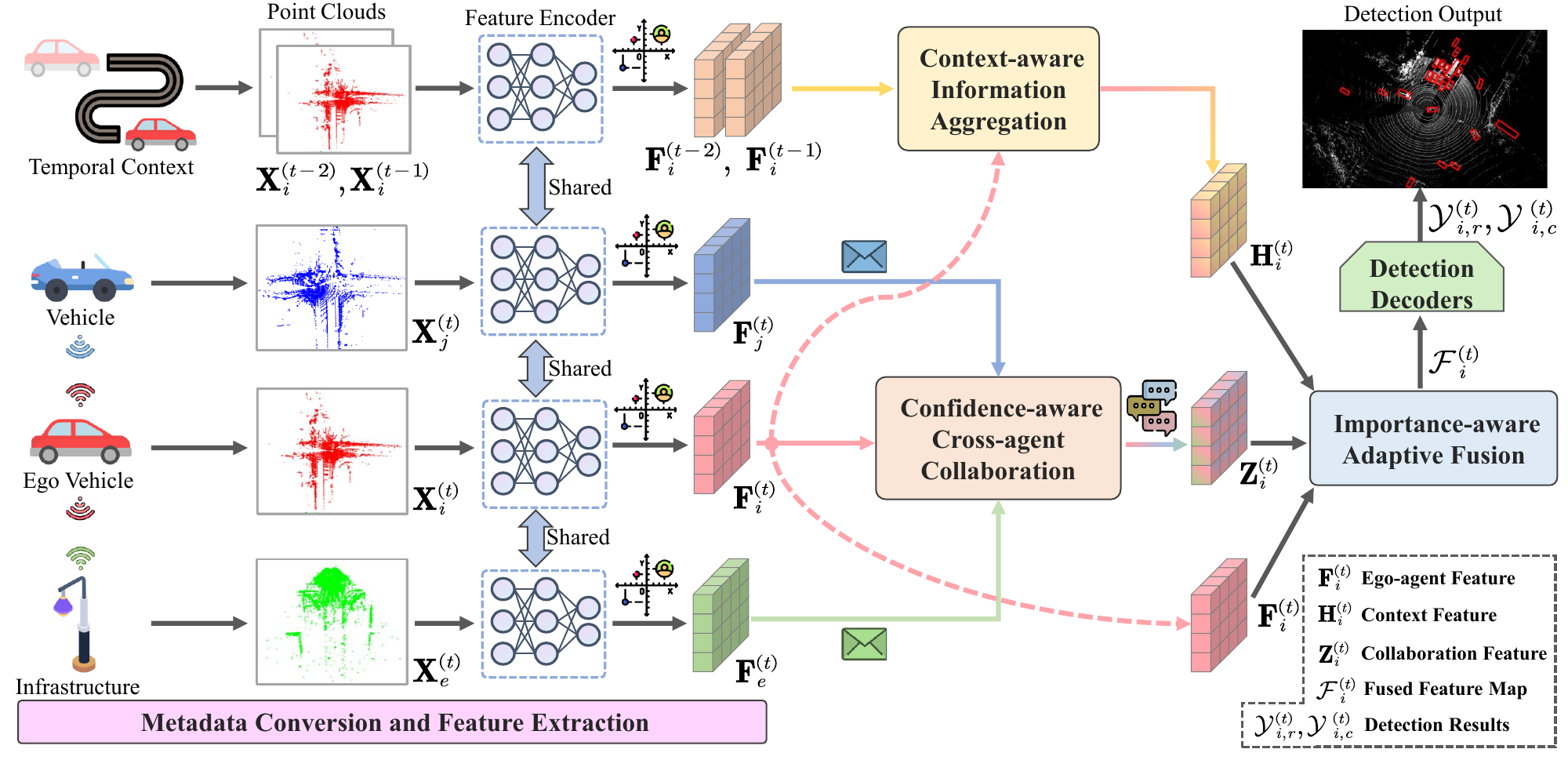}
  \caption{The overall architecture of the proposed {\tt SCOPE}. The framework consists of five parts: metadata conversion and feature extraction, context-aware information aggregation, confidence-aware cross-agent collaboration, importance-aware adaptive fusion, and detection decoders. The details of each individual component are illustrated in Section~\ref{method}.
  }
  \label{arc}
  \vspace{-10pt}
\end{figure*}

Moreover, spatial information aggregation in collaborative perception systems has exposed several problems. For the collaborator-shared message fusion, the per-agent/location-based fusion strategies of previous attempts~\cite{li2023learning,li2021learning,liu2020when2com,wang2020v2vnet,xu2022v2x} fail to deal with feature map misalignment from heterogeneous agents due to localization errors. Consequently, misleading features from collaborators could lead to object position misjudgment of the ego vehicle (\ie, receiver) and harm its detection precision.
For the ego vehicle message refinement, existing approaches~\cite{hu2022where2comm,li2021learning,liu2020when2com, luo2022complementarity,wang2020v2vnet,xu2022bridging} rely on fused representations with collaborator information to implement detection, abandoning the natural perception advantages of the ego vehicle and introducing the potential noises. The ego-centered characteristics may contain locally critical spatial information that is not disturbed by noise from collaborative agents with asynchronous sensor measurements.
To this end, how to effectively break the aforementioned limitations becomes a priority for achieving robust collaborative perception.
 
Motivated by the above observations, we present {\tt SCOPE}, a \ul{S}patio-temporal domain awareness multi-agent \ul{C}ollab\ul{O}rative \ul{PE}rception approach to tackle existing challenges jointly.
From Figure~\ref{arc}, {\tt SCOPE} achieves effective information collaboration among agents and precise perception of the ego agent via three proposed core components.
Specifically, (\rmnum{1}) for the data sparsity challenge in single-frame point clouds, the \emph{context-aware information aggregation} is proposed to aggregate the ego agent's context information from preceding frames. We employ selective information filtering and spatio-temporal feature integration modules to capture informative temporal cues and historical context semantics. (\rmnum{2}) For the collaborator-shared message fusion challenge, the \emph{confidence-aware cross-agent collaboration} is introduced to guarantee that 
the ego agent aggregates complementary information from heterogeneous agents.
Based on the confidence-aware multi-scale feature interaction, we facilitate holistic communication and mitigate feature map misalignment due to localization errors of collaborators. (\rmnum{3}) For the multi-feature integration challenge of the ego agent, the \emph{importance-aware adaptive fusion} is designed to flexibly incorporate the advantages of diverse representations based on distinctive perception contributions. 
% Benefiting from these tailored components, {\tt SCOPE} takes a meaningful step towards pragmatic and substantial collaborative perception in an end-to-end manner. 
We evaluate the superiority of {\tt SCOPE} on various collaborative 3D object detection datasets, including DAIR-V2X~\cite{yu2022dair}, OPV2V~\cite{xu2022opv2v}, and V2XSet~\cite{xu2022v2x}. Extensive experiments in real-world and simulated scenarios provide favorable evidence that our approach is competitive in multi-agent collaborative perception tasks. 

The main contributions can be summarized as follows:
\begin{itemize}
\item We present {\tt SCOPE}, a novel framework for multi-agent collaborative perception. The framework facilitates information collaboration and feature fusion among agents, achieving a reasonable 
performance-bandwidth trade-off. Comprehensive experiments on collaborative detection tasks show that {\tt SCOPE} outperforms previous state-of-the-art methods.

\item To our best knowledge, we are the first to consider the temporal context of the ego agent in collaborative perception systems. Based on the proposed context-aware component, the target frame of point clouds efficiently integrates historical cues from preceding frames to capture valuable temporal information.

\item  We introduce two spatial information aggregation components to address the challenges of collaboration heterogeneity and fused representation unicity. These tailored components effectively perform multi-grained information interactions among agents and multi-source feature refinement of the ego agent.

\end{itemize}

\section{Related Work} 
\subsection{Multi-Agent Communication}
Benefiting from rapid advances in learning-based technologies~\cite{liang2018deep,yang2023context,yang2022disentangled, yang2022contextual, yang2022learning,yang2023novel,yang2022novel}, communication has become an essential component in constructing robust multi-agent systems. Previous 
works~\cite{foerster2016learning,singh2018learning, sukhbaatar2016learning,tan1993multi} typically focus on centralized paradigms or predefined protocols to explore information-sharing mechanisms among multiple agents.
However, these efforts are potentially domain-constrained and scenario-isolated. Recent methods~\cite{das2019tarmac,hoshen2017vain,jiang2018learning} achieve effective communication strategies in diverse scenarios. For instance, Vain~\cite{hoshen2017vain} determines which agents would share information by modeling the locality of interactions. ATOC~\cite{jiang2018learning} introduces an attentional communication strategy to learn when communicating and integrating useful information to execute collaborative decisions. 
In comparison, we focus on the more challenging 3D perception tasks in complex driving scenarios. Based on the proposed confidence-aware multi-scale information interaction, {\tt SCOPE} achieves an information-critical and holistic communication pattern among heterogeneous agents.
\subsection{Collaborative Perception}
\label{CP} 
Collaborative perception investigates how to aggregate complementary perception semantics across connected agents to improve overall system performance. Current works are mainly categorized as early~\cite{chen2019cooper}, intermediate~\cite{chen2019fcooper,hu2022where2comm,li2021learning,liu2020when2com,wang2020v2vnet,xu2022cobevt,xu2022v2x,xu2022opv2v}, and late~\cite{rauch2012car2x,xu2023model} fusion, where intermediate fusion is more favoured due to its good trade-off between performance and transmission bandwidth.
In this case, several remarkable works explore efficient collaboration mechanisms.
In DiscoNet~\cite{li2021learning}, the teacher-student network is trained to aggregate the shared information.
% Meanwhile, various fusion methods are designed to enhance the information interaction across connected agents. 
V2X-ViT~\cite{xu2022v2x} proposes heterogeneous multi-agent self-attention to handle device heterogeneity, and Xu \etal~\cite{xu2022cobevt} design an axial attention module to capture sparse spatial semantics.
Furthermore, Where2comm~\cite{hu2022where2comm} filters irrelevant semantics by quantifying the spatial importance through confidence maps.
Unlike these single-frame methods, we propose an information aggregation component to extract rich context information from local historical observations. Moreover, our cross-agent collaboration mechanism outperforms the previous per-agent/location message fusion models, significantly mitigating the impact of localization errors.

\section{Methodology}
\label{method}

In this paper, we focus on developing a robust collaborative perception system to improve the ego vehicle's perception capability. 
Figure~\ref{arc} illustrates the overall architecture of the proposed framework. It consists of five sequential parts, where the system input accommodates different types of connected agents (\ie, infrastructures and AVs).

\subsection{Metadata Conversion and Feature Extraction}
During the collaboration, one of the connected agents is identified as the ego agent ($i$) to build a communication graph, where other agents within the communication range serve as collaborators. Upon receiving the metadata (\eg, poses and extrinsics) broadcast by the ego agent, 
the collaborators project their local LiDAR point clouds to the coordinate system of the ego agent. Similarly, the ego agent's previous point cloud frames are synchronized to the current frame. Each agent $k \in \{1,...,K\}$ encodes the projected point clouds into Bird's Eye View (BEV) features to extract local visual representations. Given the point cloud $\bm{X}_k^{(t)}$ of $k$-th agent at timestamp $t$, the extracted feature is $\bm{F}_{k}^{(t)} = f_{enc}(\bm{X}_{k}^{(t)}) \in \mathbb{R}^{C \times H \times W}$, where $f_{enc}(\cdot)$ is the PointPillar~\cite{lang2019pointpillars} encoder shared among all agents, and $C, H, W$ stand for the channel, height, and width.

\subsection{Context-aware Information Aggregation}
\begin{figure}[t]
  \centering
  \includegraphics[width=\linewidth]{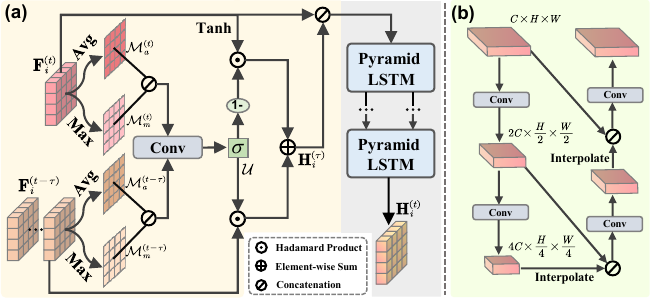}
  \caption{(a) The architecture of the proposed CIA component, including selective information filtering (\textit{left}) and spatio-temporal feature integration (\textit{middle}). (b) The multi-scale convolutional structure in the pyramid LSTM.
  }
  \label{temporal}
  \vspace{-9pt}
\end{figure}

Conventional 3D detection tasks~\cite{yuan2021temporal,zhou2022centerformer} have shown the potential of temporal information in improving perception performance. Unfortunately, current multi-agent collaborative efforts~\cite{hu2022where2comm,li2021learning,liu2020when2com,luo2022complementarity,wang2020v2vnet} invariably focus on exploring the current frame while ignoring context cues from previous frames. The single-frame solution fails to effectively detect fast-moving objects (\eg, surrounding vehicles) due to the sparsity and insufficiency of point clouds. To this end, we present a Context-aware Information Aggregation (CIA) component to capture spatio-temporal representations of the ego agent's previous frames to fuse valuable semantics. As shown in Figure~\ref{temporal}(a), CIA consists of two core phases.

\noindent\textbf{Selective Information Filtering.}
 This phase aims to extract refined information by filtering redundant features from the target $\bm{F}_{i}^{(t)} \in \mathbb{R}^{ C \times H \times W}$ and distilling meaningful characteristics from the previous $\bm{F}_{i}^{(t-\tau)} \in \mathbb{R}^{\tau \times C \times H \times W}$, where $\tau$ is the time shift. Concretely, the channel-wise average and max pooling operations $\Psi_{a/m}(\cdot)$ are first utilized to aggregate rich channel semantics, denoted as $\mathcal{M}_{a/m}^{(t)}=\Psi_{a/m}(\bm{F}_{i}^{(t)}) $ and $\mathcal{M}_{a/m}^{(\tau)}=\Psi_{a/m}(\bm{F}_{i}^{(\tau)})$, where $\bm{F}_{i}^{(\tau)} =  {\textstyle \sum_{n=1}^{\tau}}\bm{F}_{i}^{(t-n)}$. These refined feature maps are combined to obtain a well-chosen spatial selection map. The procedure is expressed as follows:
\begin{equation}
    \mathcal{U} = \sigma \cdot \aleph(\mathcal{M}_{a}^{(t)} \Vert \mathcal{M}_{m}^{(t)} + \mathcal{M}_{a}^{(\tau)} \Vert \mathcal{M}_{m}^{(\tau)}) \in \mathbb{R}^{H \times W}, 
\end{equation}
where $\sigma$ is the sigmoid activation, $\aleph$ denotes a convolutional layer for fusion, and $\Vert$ denotes the concatenation operation. Immediately, the information filtering strategy is implemented via pixel-wise attention weighting to enhance the salient historical features of $\bm{H}_{i}^{(\tau)} \in \mathbb{R}^{\tau \times C \times H \times W}$:
\begin{equation}
    \bm{H}_{i}^{(\tau)} = (1-\mathcal{U}) \odot  tanh(\bm{F}_{i}^{(t)}) + \mathcal{U} \odot \bm{F}_{i}^{(t-\tau)}.
\end{equation}

\noindent\textbf{Spatio-temporal Feature Integration.}
In order to integrate the historical prototype to improve the perception ability of the current representation, we introduce a pyramid LSTM~\cite{lei2022latency} to learn the contextual dependencies of inter-frame features $\bm{F}_{i}^{(t)} \Vert \bm{H}_{i}^{(\tau)}$. The difference between the pyramid LSTM and vanilla LSTM~\cite{hochreiter1997long} is that the initial matrix multiplication is replaced with a multi-scale convolutional structure, as shown in Figure~\ref{temporal}(b). Note that the intrinsic long short-term memory mechanism in the LSTM is adopted, and please refer to~\cite{hochreiter1997long} for details. In practice, multi-scale spatial features are extracted by two successive 2D convolutions among different scales, followed by batch normalization and ReLU activation. To achieve multi-level spatial semantic fusion, the downsampled features are progressively interpolated to deliver to the upsampled layers via lateral connections. The overall design facilitates learning spatio-temporal correlations in historical contexts, which benefits our multi-frame perception paradigm. Ultimately, the final hidden state in the pyramid LSTM is used as the refined context-aware feature $\bm{H}_{i}^{(t)} \in \mathbb{R}^{ C \times H \times W}$.

 \begin{figure}[t]
  \centering
  \includegraphics[width=\linewidth]{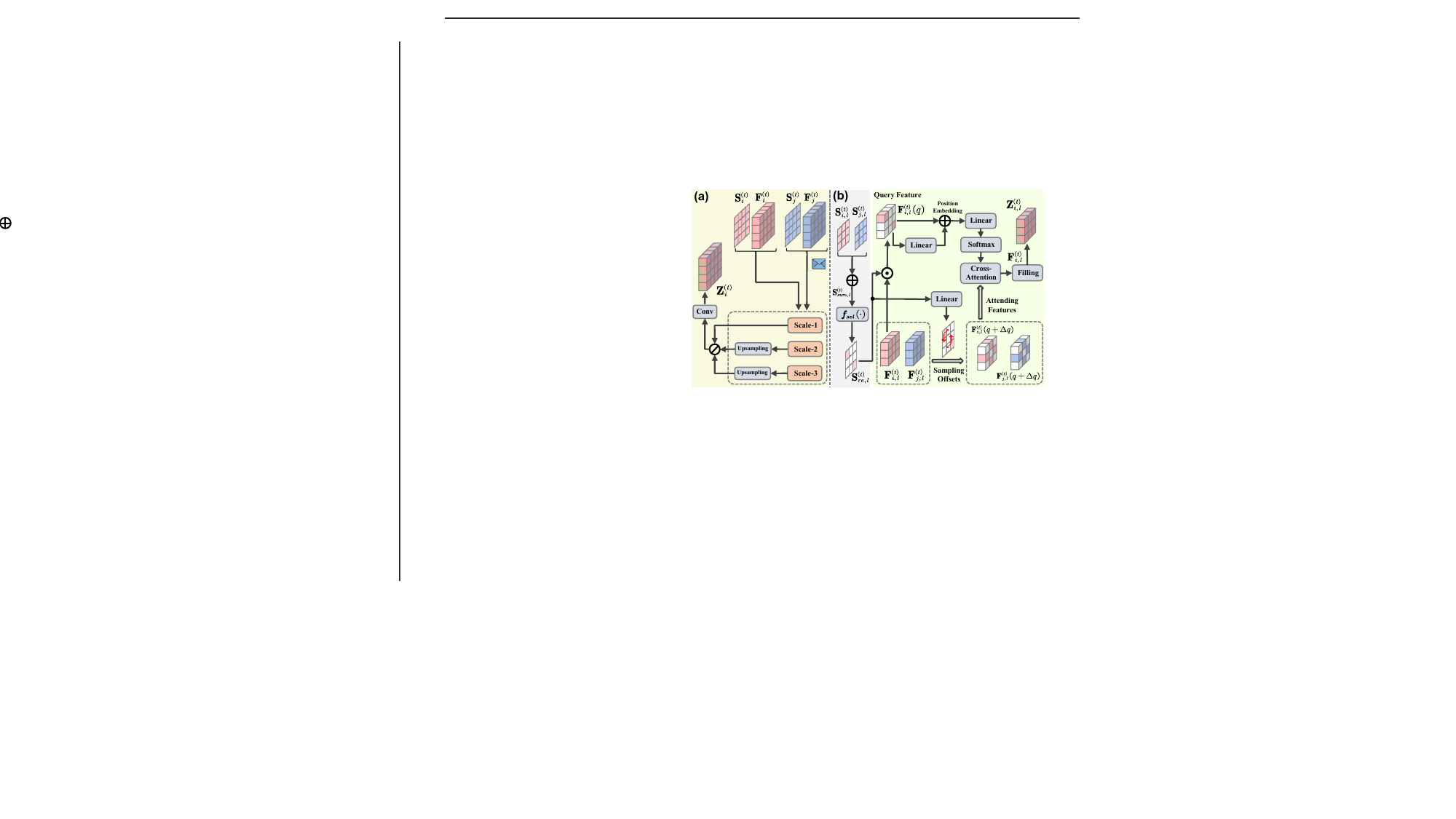}
  \caption{(a) The architecture of the proposed CCC component. (b) CCC consists of two phases at each scale: reference point proposal (\emph{middle}) and deformable cross-attention module (\emph{right).}
  }
  \label{colla}
  \vspace{-8pt}
\end{figure}
  
\subsection{Confidence-aware Cross-agent Collaboration}
The cross-agent collaboration targets to enhance the ego agent's visual representations by aggregating the complementary semantics from the collaborator-shared features. 
Existing works~\cite{hu2022where2comm,hu2023collaboration,xu2022v2x,xu2022opv2v} propose attention-based per-location feature fusion methods, which are vulnerable to localization errors and ignore the sparsity of point clouds. To tackle these issues, we implement a novel Confidence-aware Cross-agent Collaboration (CCC) component.
We first obtain the confidence map $\bm{S}_{k}^{(t)}$ and binary selection matrix $\mathcal{C}_{k}^{(t)}$ of feature $\bm{F}_{k}^{(t)}, k\in \{1,...,K\}$ following~\cite{hu2022where2comm}. The filtered feature shared by $k$-th agent to ego agent is $\bm{F}_{k}^{(t)} = \bm{F}_{k}^{(t)} \odot \mathcal{C}_{k}^{(t)}$. 
As Figure~\ref{colla}(a) shows, the CCC component encodes the features and confidence maps into three scales and performs feature fusion at each scale. $\bm{F}_{k,l}^{(t)}$ and $\bm{S}_{k,l}^{(t)}$ denote the feature and confidence map at $l$-th scale.
We take two agents as an example to elaborate on the feature fusion in Figure~\ref{colla}(b), including the following two phases.

\noindent\textbf{Reference Point Proposal.}
To predict the spatial locations containing meaningful object information, we obtain the element-wise summation of all confidence maps as $\bm{S}_{sum,l}^{(t)} = \sum_{k=1}^{K} \bm{S}_{k,l}^{(t)}$. Since the confidence map reflects the spatially critical levels, $\bm{S}_{sum,l}^{(t)}$ presents the potential positions of the targets within the detection range, called the reference points. Accordingly, we apply a threshold-based selection function $f_{sel}(\cdot)$ to extract the reference points $\bm{S}_{re,l}^{(t)}$.
This design could actively guide the subsequent fusion network to concentrate on the essential spatial areas.

\noindent\textbf{Deformable Cross-attention Module.}
We extract the ego agent's feature $\bm{F}_{i,l}^{(t)}$ at the reference points as the initial query embedding and apply a linear layer to encode the locations of reference points into a position embedding. 
To resolve the misalignment of feature maps and acquire more robust representations against localization errors, the Deformable Cross-attention Module (DCM) aggregates the information from sampled keypoints through a deformable cross-attention layer~\cite{zhu2021deformable}. Concretely,
a linear layer is leveraged to learn the 2D spatial offset $\Delta q$ of the reference point $q$ for sampling attending keypoints at $q + \Delta q$. We simultaneously select $M$ keypoints and extract their features as the attending features. The output of DCM at position $q$ is:
\begin{equation}
\footnotesize
    \text{DCM}(q) = \sum_{a=1}^{A} \bm{W}_a [\sum_{k=1}^K \sum_{m=1}^M \phi(\bm{W}_b \bm{F}_{i,l}^{(t)}(q)) \bm{F}_{k,l}^{(t)}(q+\Delta q_m)],
\end{equation}
where $a$ indexes the attention head, $\bm{W}_{a/b}$ are the learnable weights, and $\phi(\cdot)$ is the softmax function.
Ultimately, we derive a filling operation that fills $\text{DCM}(q)$ into the ego agent's feature $\bm{F}_{i,l}^{(t)}$ based on the initial position $q$ and outputs $\bm{Z}_{i,l}^{(t)}$.
The output-enhanced features under the three scales are encoded into the same size and concatenated in the channel dimension. We utilize a $1 \times 1$ convolutional layer to fuse information of the three scales and get the final collaboration feature $\bm{Z}_{i}^{(t)} \in \mathbb{R}^{C \times H \times W}$.

\subsection{Importance-aware Adaptive Fusion}
 \begin{figure}[t]
  \centering
  \includegraphics[width=\linewidth]{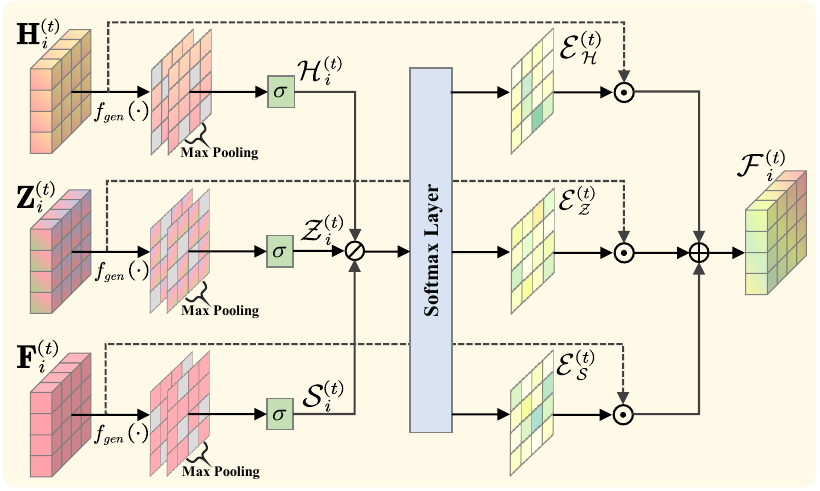}
  \caption{The architecture of the proposed IAF component.
  }
  \label{fusion}
  \vspace{-8pt}
\end{figure}

Although previous works~\cite{li2021learning,liu2020when2com,luo2022complementarity,wang2020v2vnet,xu2022v2x} have achieved impressive performance through ego features that aggregate collaborative information, they potentially suffer from noise interference introduced by collaborators with asynchronous measurements. 
A promising solution is to consider purely ego-centered characteristics, which contain the natural perception advantages of the target agent.
To this end, we propose an Importance-aware Adaptive Fusion (IAF) component to fuse multi-source features based on their complementary contributions. These informative features include $\bm{H}_{i}^{(t)}, \bm{Z}_{i}^{(t)}$, and $\bm{F}_{i}^{(t)}$. 
From Figure~\ref{fusion}, the importance generator $f_{gen}(\cdot)$ is used to generate the spatial importance maps, which processes as follows:
\begin{equation}
\mathcal{H}_{i}^{(t)},\mathcal{Z}_{i}^{(t)},\mathcal{S}_{i}^{(t)} = \sigma \cdot
\Psi_{m} (f_{gen}(\bm{H}_{i}^{(t)}, \bm{Z}_{i}^{(t)}, \bm{F}_{i}^{(t)})),
\end{equation}
where $\mathcal{H}_{i}^{(t)},\mathcal{Z}_{i}^{(t)},\mathcal{S}_{i}^{(t)} \in [0,1]^{H \times W}$. Then, these importance maps are concatenated to get the normalized attention maps across distinct features by the softmax function $\phi(\cdot)$:
\begin{equation}
\small
\mathcal{E}_{\mathcal{H}}^{(t)}  = \phi(\mathcal{H}_{i}^{(t)}) = \frac{exp(\mathcal{H}_{i}^{(t)})}{exp(\mathcal{H}_{i}^{(t)})+exp(\mathcal{Z}_{i}^{(t)})+exp(\mathcal{S}_{i}^{(t)})}.
\end{equation}
A larger value in $\mathcal{E}_{\mathcal{H}}^{(t)} \in \mathbb{R}^{H \times W}$ implies that the corresponding feature position is more significant.
% The larger $\mathcal{E}_{\mathcal{H}}^{(t)}$ implies that the corresponding feature position is more important. 
Similarly, $\mathcal{E}_{\mathcal{Z}}^{(t)}  = \phi(\mathcal{Z}_{i}^{(t)})$ and
$\mathcal{E}_{\mathcal{S}}^{(t)}  = \phi(\mathcal{S}_{i}^{(t)})$. Finally, we adaptively activate each feature map's perceptually critical spatial information by a weighted summation. The fused feature map $\mathcal{F}_{i}^{(t)} \in \mathbb{R}^{C \times H \times W}$ is obtained as follows:
\begin{equation}
\mathcal{F}_{i}^{(t)} = \mathcal{E}_{\mathcal{H}}^{(t)} \odot  \bm{H}_{i}^{(t)} + \mathcal{E}_{\mathcal{Z}}^{(t)} \odot  \bm{Z}_{i}^{(t)} + \mathcal{E}_{\mathcal{S}}^{(t)} \odot \bm{F}_{i}^{(t)}.
\end{equation}
 
\subsection{Detection Decoder and Objective Optimization}
\label{dd}
We leverage two detection decoders to decode the final fused feature $\mathcal{F}_{i}^{(t)}$ into the predicted outputs. The regression output is $\mathcal{Y}_{i,r}^{(t)} = f_{dec}^r(\mathcal{F}_{i}^{(t)}) \in \mathbb{R}^{7 \times H \times W}$, representing the position, size, and yaw angle of the bounding box at each location. The classification output reveals the confidence value of each predefined box to be a target or background, which is $\mathcal{Y}_{i,c}^{(t)} = f_{dec}^c(\mathcal{F}_{i}^{(t)}) \in \mathbb{R}^{2 \times H \times W}$. Following existing work~\cite{lang2019pointpillars}, we adopt the smooth $L_1$ loss for regressing the bounding boxes and a focal loss~\cite{lin2017focal} for classification.

\section{Experiments}
\subsection{Datasets and Evaluation Metrics}
\noindent\textbf{Datasets.} We conduct comprehensive experiments on three benchmark datasets for collaborative perception tasks, including DAIR-V2X~\cite{yu2022dair}, V2XSet~\cite{xu2022v2x}, and OPV2V~\cite{xu2022opv2v}. These datasets provide annotated sensor data from multiple connected agents for each sample. 
Concretely, \textbf{DAIR-V2X} is a large-scale real-world dataset supporting collaborative 3D object detection, containing 71,254 samples and dividing the training/validation/testing sets as 5:2:3. Each sample contains the LiDAR point clouds from a vehicle and an infrastructure. During feature extraction, we transform the point cloud into the BEV feature map of the size (200, 504, 64). 
\textbf{V2XSet} is a public simulation dataset for V2X collaborative perception, including 73 representative scenarios and 11,447 frames of labeled point clouds generated by Carla~\cite{dosovitskiy2017carla}. The training/validation/testing sets are 6,694, 1,920, and 2,833 frames, respectively. The size of each extracted feature map is (100, 384, 64). 
\textbf{OPV2V} is a vehicle-to-vehicle collaborative perception dataset collected by Carla and OpenCDA~\cite{xu2021opencda}. It includes 11,464 frames of point clouds and RGB images with 3D annotation, splitting into training/validation/testing sets of 6,764, 1,981, and 2,719 frames, respectively. The 3D point cloud is extracted as a feature map of the size (100, 384, 64).

\noindent\textbf{Evaluation Metrics.} Following~\cite{xu2022v2x}, we leverage the Average Precision (AP) at Intersection-over-Union (IoU) thresholds of 0.5 and 0.7 to evaluate the 3D object detection performance. The communication results count the message size by byte in the log scale with base 2~\cite{hu2022where2comm}.

\subsection{Implementation Details.} We build the proposed models on the Pytorch toolbox~\cite{paszke2019pytorch} and train them on four Nvidia Tesla V100 GPUs using the Adam optimizer~\cite{kingma2015adam}. The training settings of the DAIR-V2X, V2XSet, and OPV2V datasets follows: the batch sizes are \{5, 3, 3\} and the epochs are \{30, 40, 40\}. The initial learning rate is 2e-3 and decays every 15 epochs with a factor of 0.1. Our CIA module uses the previous 2 frames, the attention head in the CCC component is 8, and the number of keypoints is 15.
To simulate the default localization and heading errors, we add Gaussian noise with a standard deviation of 0.2m and 0.2$^{\circ}$ to the initial position and orientation of the agents. The voxel resolution of the observation encoder $f_{enc}(\cdot)$ is set to 0.4m in height and width. The generator $f_{gen}(\cdot)$ follows a detection decoder from~\cite{lang2019pointpillars}, and the summed confidence map $\bm{S}_{sum,l}^{(t)}$ is filtered using a threshold of 0.1 in the function $f_{sel}(\cdot)$. The decoders $\{f_{dec}^r(\cdot), f_{dec}^c(\cdot)\}$ are derived with two 1 $\times$ 1 convolutional layers. We provide results on the validation set of DAIR-V2X and the testing set of V2XSet and OPV2V.

\begin{table}[t]
\caption{Performance comparison on the DAIR-V2X, V2XSet, and OPV2V datasets. The results are reported in AP@0.5/0.7.}
\centering
\resizebox{\linewidth}{!}{%
\begin{tabular}{c|c|c|c}
\toprule
\multirow{2}{*}{Model} & DAIR-V2X    & V2XSet     & OPV2V       \\ \cline{2-4} 
                       & AP@0.5/0.7   & AP@0.5/0.7   & AP@0.5/0.7   \\ \midrule
No Fusion       & 50.03/43.57 & 60.60/40.20 & 68.71/48.66 \\
Late Fusion            & 53.12/37.88 & 66.79/50.95 & 82.24/65.78 \\
Early Fusion            & 61.74/46.53 & 77.39/50.45 & 81.30/68.69 \\
When2com~\cite{liu2020when2com}                & 51.12/36.17 & 70.16/53.72 & 77.85/62.40 \\
V2VNet~\cite{wang2020v2vnet}                  & 56.01/42.25 & 81.80/61.35 & 82.79/70.31 \\
AttFuse~\cite{xu2022opv2v}                & 53.79/42.61 & 76.27/57.93 & 83.21/70.09 \\
V2X-ViT~\cite{xu2022v2x}                & 54.26/43.35 & 85.13/68.67 & 86.72/74.94 \\
DiscoNet~\cite{li2021learning}                & 54.29/44.88 & 82.18/63.73 & 87.38/73.19 \\
CoBEVT~\cite{xu2022cobevt}                & 54.82/43.95 & 83.01/62.67 & 87.40/74.35 \\
Where2comm~\cite{hu2022where2comm}              & 63.71/48.93 & 85.78/72.42 & 88.07/75.06 \\
\textbf{SCOPE (Ours)}      & \textbf{65.18/49.89} & \textbf{87.52/75.05} & \textbf{89.71/80.62} \\ \bottomrule
\end{tabular}
}
\label{tabcom}
\vspace{-0.3cm}
\end{table}

\begin{figure*}[t]
    \centering
    % \centerline{\includesvg[width=1\textwidth]{result/comm_vol.svg}}
    \centerline{\includegraphics[width=1\textwidth]{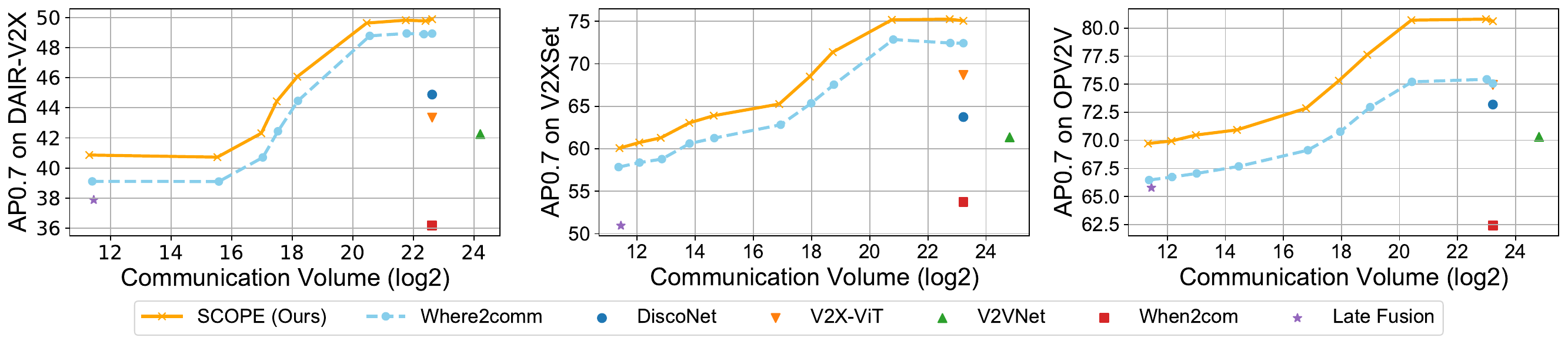}}
    % \includegraphics[width=\linewidth]{imgs/fusion.pdf}
    % \vspace{-0.5cm}
    \caption{Collaborative perception performance comparison of {\tt SCOPE} and Where2comm~\cite{hu2022where2comm} on the DAIR-V2X, V2XSet and OPV2V datasets with varying communication volume.}
    \label{fig_comm}
    \vspace{-0.2cm}
\end{figure*}

\begin{figure*}[h]
    \centering
    % \centerline{\includesvg[width=1\textwidth]{result/local_error.svg}}
    \centerline{\includegraphics[width=1\textwidth]{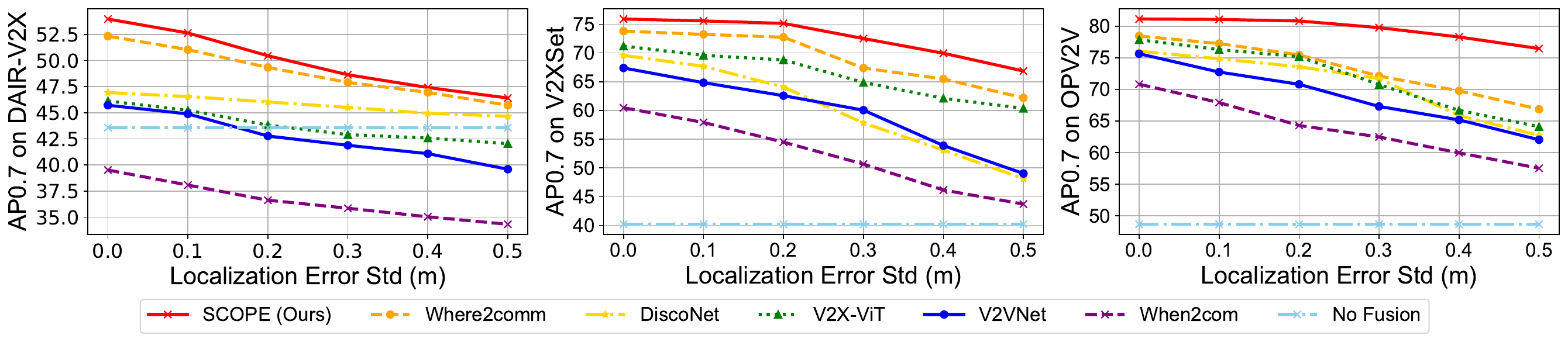}}
    \caption{Robustness to the localization error on the DAIR-V2X, V2XSet, and OPV2V datasets.}
    \label{fig_loc}
    \vspace{-0.2cm}
\end{figure*}

\begin{figure*}[t]
    \centering
    % \centerline{\includesvg[width=1\textwidth]{result/head_error.svg}}
    \centerline{\includegraphics[width=1\textwidth]{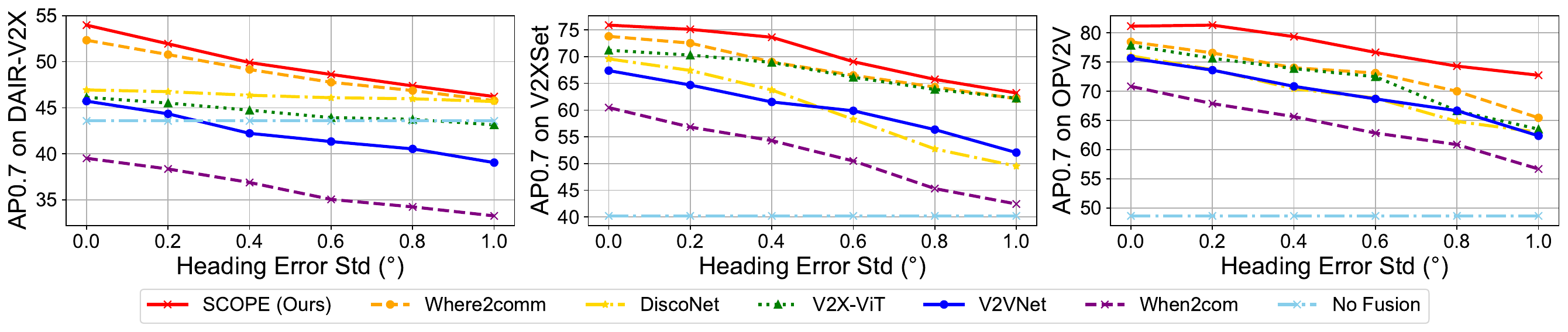}}
    % \vspace{-0.5cm}
    \caption{Robustness to the heading error on the DAIR-V2X, V2XSet, and OPV2V datasets.}
    \label{fig_head}
    \vspace{-0.2cm}
\end{figure*}

\subsection{Quantitative Evaluation}
\noindent\textbf{Detection Performance Comparison.}
Table~\ref{tabcom} shows the 3D detection performance comparison results on the three datasets. We compare the following methods. No Fusion as a baseline considers only the point cloud data of ego vehicles without collaboration. 
Late Fusion allows agents to exchange detected results and utilizes non-maximum suppression to produce the final outputs.
Early Fusion means the raw LiDAR point clouds are shared to complete the perception view.
Meanwhile, the existing state-of-the-art (SOTA) models are fully considered, including When2com~\cite{liu2020when2com}, V2VNet~\cite{wang2020v2vnet}, AttFuse~\cite{xu2022opv2v}, V2X-ViT~\cite{xu2022v2x}, DiscoNet~\cite{li2021learning}, CoBEVT~\cite{xu2022cobevt}, and Where2comm~\cite{hu2022where2comm}.
Intuitively, the proposed {\tt SCOPE} outperforms No Fusion, Late Fusion, and Early Fusion by large margins across all datasets, demonstrating the superiority of our perception paradigm. 
Moreover, our method significantly outperforms previous SOTA models in simulated and real-world (DAIR-V2X) scenarios.
In particular, the SOTA performance is improved by 3.63\% and 7.41\% on the V2XSet and OPV2V datasets separately in AP@0.7.
Compared to single-frame and collaboration-only based efforts~\cite{hu2022where2comm,li2021learning,liu2020when2com,wang2020v2vnet,xu2022v2x}, {\tt SCOPE} proves the effectiveness and rationality of simultaneously considering the temporal context, collaboration, and ego characteristics.

\noindent\textbf{Comparison of Communication Volume.}
Evaluating performance at varying bandwidths plays an essential role in achieving pragmatic perception. To this purpose, Figure~\ref{fig_comm} shows the perception performance of our {\tt SCOPE} and Where2comm~\cite{hu2022where2comm} with various communication volumes. 
Note that we do not consider communication methods that use any extra model/data/feature compression for a fair comparison. We have the following observations:
(i) {\tt SCOPE} achieves better performance than previous models~\cite{li2021learning,liu2020when2com,wang2020v2vnet,xu2022v2x} on all datasets with less communication volume.
(ii) Compared to Where2comm~\cite{hu2022where2comm}, our method obtains a superior perception-communication trade-off across all the communication bandwidth choices.

\noindent\textbf{Robustness to Localization and Heading Errors.}
Following the noise settings from \cite{xu2022v2x}, we perform experiments to evaluate the robustness against natural localization noises of existing methods.
The noises are sampled from a Gaussian distribution with standard deviation $\sigma_{xyz} \in [0, 0.5]$ m, $\sigma_{\text{heading}} \in [0^{\circ}, 1.0^{\circ} ]$. From Figures~\ref{fig_loc} and \ref{fig_head}, the overall collaborative perception performance consistently deteriorates as errors continue to increase.
For instance, When2com and V2VNet fail even worse than No Fusion on the DAIR-V2X when the localization and heading errors are over 0.2 m and 0.4$^{\circ}$, respectively. Noticeably, our method consistently surpasses the previous SOTA models under all noise levels, clearly displaying the robustness of {\tt SCOPE} against pose errors. The reasonable explanations are: (i) the deformable cross-attention module facilitates capturing perceptually critical information across agents; (ii) the proposed multi-scale feature interaction alleviates the feature map misalignment about collaborative agents due to noises.

\begin{table}[t]
\caption{Ablation study results of the proposed core components on all the datasets. \textbf{PL}: pyramid LSTM; \textbf{SIF}: selective information filtering; \textbf{DCM}: deformable cross-attention module; \textbf{RPP}: reference point proposal; \textbf{IAF}: importance-aware adaptive fusion.}
\centering
\resizebox{\linewidth}{!}{%
\begin{tabular}{ccccc|c|c|c}
\toprule
\multirow{2}{*}{PL} & \multirow{2}{*}{SIF} & \multirow{2}{*}{DCM} & \multirow{2}{*}{RPP} & \multirow{2}{*}{IAF} & DAIR-V2X    & V2XSet     & OPV2V       \\ \cline{6-8} 
                     &                     &                      &                      &                      & AP@0.5/0.7 & AP@0.5/0.7 & AP@0.5/0.7 \\ \midrule
                     &                     &                      &                      &                      & 60.31/40.35       & 76.04/62.41       & 77.22/65.51       \\
\CheckmarkBold                    &                     &                      &                      &                      & 61.75/44.58       & 80.36/65.74       & 81.19/69.02       \\
\CheckmarkBold                      & \CheckmarkBold                     &                      &                      &                      & 62.48/45.62       & 81.88/67.42       & 82.34/71.23    \\
\CheckmarkBold                      & \CheckmarkBold                     & \CheckmarkBold                      &                      &                      & 64.06/47.87       & 85.47/72.29       & 86.96/76.84       \\
\CheckmarkBold                     & \CheckmarkBold                     & \CheckmarkBold                     & \CheckmarkBold                     &                      & 64.74/48.45       & 86.33/73.27       & 88.01/78.57       \\
\CheckmarkBold                      & \CheckmarkBold                    &\CheckmarkBold                      & \CheckmarkBold                      & \CheckmarkBold                      & \textbf{65.18/49.89}      & \textbf{87.52/75.05}       & \textbf{89.71/80.62}       \\ \bottomrule
\end{tabular}
}
\label{tab_ab1}
\vspace{-0.3cm}
\end{table}
\subsection{Ablation Studies}
We perform thorough ablation studies on all datasets to understand the necessity of the different designs.
Tables~\ref{tab_ab1} and  \ref{tab_ab2} show the following observations.

\noindent\textbf{Importance of Core Components.} We investigate the contribution of each core component in {\tt SCOPE}. A base version (first row of Table~\ref{tab_ab1}) is provided as the gain reference. In the base model, we use the same fusion pattern to aggregate context information, collaborator-shared features, and multi-source features, where features are concatenated on the channel dimension and fused by a 1 $\times$ 1 convolutional network. As Table~\ref{tab_ab1} demonstrates, we progressively add (1) Pyramid LSTM, (2) SIF, (3) DCM, (4) RPP, and (5) IAF and present the corresponding detection precision. The consistently increased detection results over the three datasets reveal the effectiveness of each proposed component, which all provide valuable performance gains.

\noindent\textbf{Effect of Historical Frame Number.}
As shown in the top half of Table~\ref{tab_ab2}, we evaluate the effect of integrating different degrees of the temporal context from the ego agent. The interesting finding is that choosing a suitable historical frame count (\ie, 2 frames) is beneficial in achieving better performance gains. The potential reason is that adding excessive historical frames may lead to accumulated errors under the noise setting and degrade the performance.

\begin{table}[t]
\caption{Ablation study results of candidate designs and strategies on all the datasets. ``w/'' and ``w/o'' mean the with and without.}
\centering
\resizebox{\linewidth}{!}{%
\begin{tabular}{cccc} 
\toprule
\multicolumn{1}{c|}{\multirow{2}{*}{Designs/Strategies}}     & \multicolumn{1}{c|}{DAIR-V2X}    & \multicolumn{1}{c|}{V2XSet}   & OPV2V       \\ \cline{2-4}  
\multicolumn{1}{c|}{}                           & \multicolumn{1}{c|}{AP@0.5/0.7} & \multicolumn{1}{c|}{AP@0.5/0.7} & AP@0.5/0.7 \\ \midrule
\multicolumn{1}{c|}{\textbf{Full Model}}                    & \multicolumn{1}{c|}{\textbf{65.18/49.89}}       & \multicolumn{1}{c|}{\textbf{87.52/75.05}}       & \textbf{89.71/80.62}       \\ \midrule
\multicolumn{4}{c}{Effect of Historical Frame Number}                                                                                   \\ \midrule
\multicolumn{1}{c|}{1 Frame}                    & \multicolumn{1}{c|}{64.28/49.16}       & \multicolumn{1}{c|}{86.16/74.21}       & 88.85/78.49       \\
\multicolumn{1}{c|}{2   Frames (Default)}         & \multicolumn{1}{c|}{65.18/49.89}       & \multicolumn{1}{c|}{87.52/75.05}       & 89.71/80.62       \\
\multicolumn{1}{c|}{3 Frames}                    & \multicolumn{1}{c|}{65.03/49.24}       & \multicolumn{1}{c|}{86.99/74.65}       & 89.67/80.45       \\ \hline
\multicolumn{4}{c}{Effect of Keypoint Number}                                                                                    \\ \midrule
\multicolumn{1}{c|}{9 Keypoints}                & \multicolumn{1}{c|}{64.36/49.22}       & \multicolumn{1}{c|}{86.34/74.07}       & 88.32/77.94       \\
\multicolumn{1}{c|}{15   Keypoints (Default)}    & \multicolumn{1}{c|}{65.18/49.89}       & \multicolumn{1}{c|}{87.52/75.05}       & 89.71/80.62       \\
\multicolumn{1}{c|}{21   Keypoints}             & \multicolumn{1}{c|}{64.72/49.40}       & \multicolumn{1}{c|}{87.12/74.63}       & 89.08/78.38       \\ \hline
\multicolumn{4}{c}{Importance of Distinct Features}                                                                                 \\ \midrule
\multicolumn{1}{c|}{w/o  Ego-agent Featue}      & \multicolumn{1}{c|}{64.96/49.34}       & \multicolumn{1}{c|}{86.47/74.24}       & 88.63/79.38       \\
\multicolumn{1}{c|}{w/o   Context Feature}       & \multicolumn{1}{c|}{64.14/48.81}       & \multicolumn{1}{c|}{85.93/73.88}       & 88.26/77.18       \\
\multicolumn{1}{c|}{w/o   Collaboration Feature} & \multicolumn{1}{c|}{52.35/44.46}       & \multicolumn{1}{c|}{63.91/45.34}       & 71.36/50.94       \\ \hline
\multicolumn{4}{c}{Effect  of Fusion Strategies}                                                                                   \\ \midrule
\multicolumn{1}{c|}{w/ Summation Fusion}                  & \multicolumn{1}{c|}{63.65/48.91}       & \multicolumn{1}{c|}{85.81/72.64}       & 88.05/76.13       \\
\multicolumn{1}{c|}{w/ Maximum Fusion}                    & \multicolumn{1}{c|}{64.92/49.27}       & \multicolumn{1}{c|}{86.28/73.61}       & 87.95/78.17      \\
\multicolumn{1}{c|}{w/ Average Fusion}              & \multicolumn{1}{c|}{64.85/48.96}       & \multicolumn{1}{c|}{86.55/73.46}       & 88.24/78.73       \\ \bottomrule
\end{tabular}
}
\label{tab_ab2}
\vspace{-0.3cm}
\end{table}

\noindent\textbf{Effect of Keypoint Number.} We investigate the rationality of the keypoint number in the deformable cross-attention. 
Noticeably, the DCM component with 15 keypoints achieves better detection performance than other settings. This result shows that {\tt SCOPE} mitigates the impact of localization errors and fully extracts essential spatial semantics by setting a reasonable keypoint number.

\noindent\textbf{Importance of Distinct Features.}
We explore the importance of distinct representations by removing the
ego-agent, context, and collaboration features (\ie, $\bm{F}_{i}^{(t)}$, $\bm{H}_{i}^{(t)}$, $\bm{Z}_{i}^{(t)}$). There is a significant drop in the model's performance when the collaboration feature is removed, indicating that multi-agent information sharing dominates various collaborative perception tasks. Furthermore, the consistently decreased performance suggests that ego-centered characteristics (\ie, pure ego-agent feature and corresponding context information) provide an indispensable contribution.

 \begin{figure*}[t]
  \centering
  \includegraphics[width=0.9\linewidth]{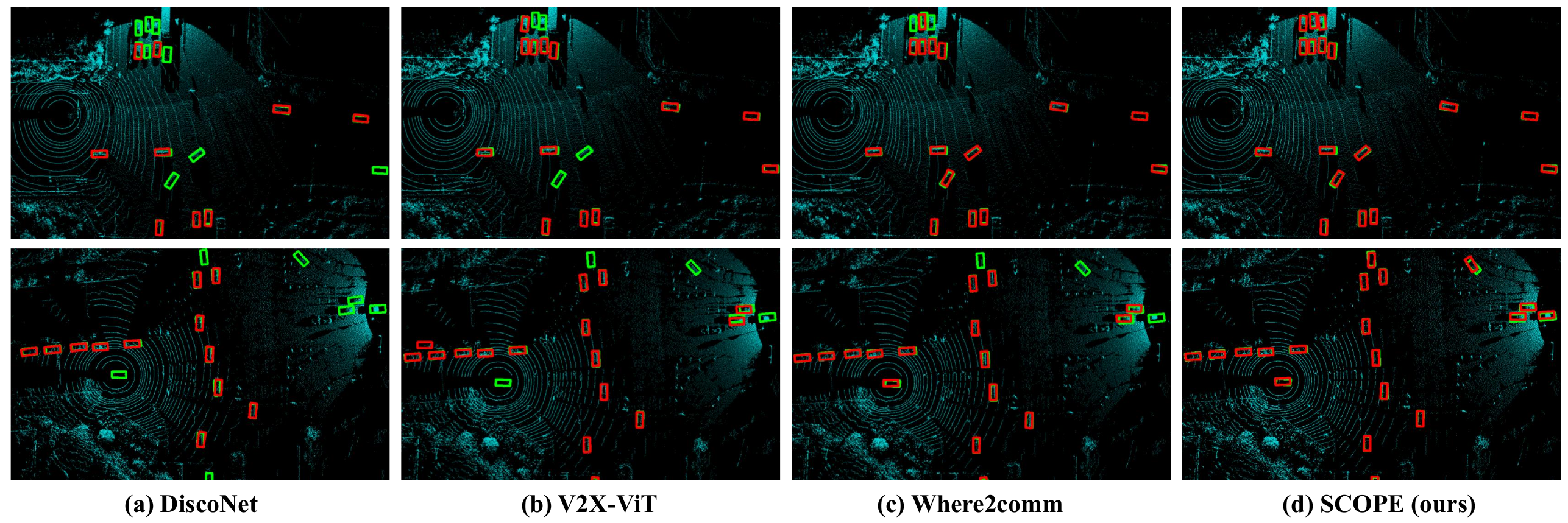}
  \caption{Qualitative comparison results in real-world scenarios from the DAIR-V2X dataset. \textcolor{green}{Green} and \textcolor{red}{red} boxes denote ground truths and detection results, respectively. Compared to the previous SOTA models, our method achieves more accurate detection results.
  }
  \label{vis_det}
  \vspace{-5pt}
\end{figure*}

 \begin{figure*}[t]
  \centering
  \includegraphics[width=0.9\linewidth]{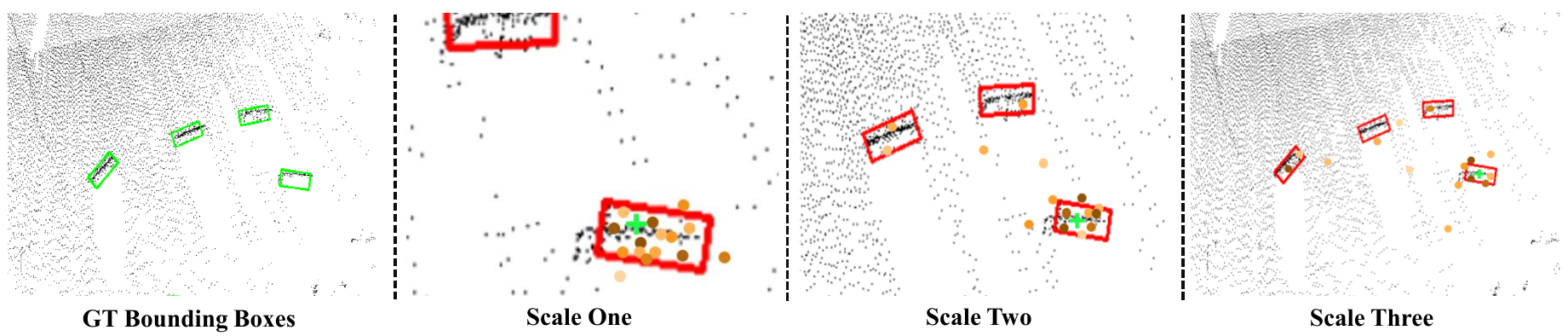}
  \caption{Visualization of learned multi-scale deformable cross-attention. \textcolor{green}{Green} cross marker denotes the proposed reference point. The sampled keypoints are represented by the \textcolor{brown}{brown} dots whose colors reflect the corresponding attention weights.
  }
  \label{vis_attention}
  \vspace{-5pt}
\end{figure*}

\noindent\textbf{Effect of Fusion Strategies.}
Finally, we demonstrate the effectiveness of the proposed adaptive fusion strategy by providing candidate fusion mechanisms.
Summation Fusion is a direct pixel-wise addition operation. Maximum and Average Fusion select elements at corresponding positions among features to retain the maximum or average, respectively, and then reshape them into a fused representation. The poor results in Table~\ref{tab_ab2} show that the above mechanisms fail to tackle the complex multi-source feature fusion challenge. In contrast, the superiority of our IAF module is reflected in considering the complementary contributions of each feature at different positions, adaptively assigning higher importance to spatially beneficial semantics.

\subsection{Qualitative Evaluation}
\noindent\textbf{Visualization of Detection Results.}
To intuitively illustrate the perception capability of different models, we provide a detection visualization comparison in Figure~\ref{vis_det} under two challenging scenarios. {\tt SCOPE} achieves more accurate and robust detection results than previous SOTA methods, where the predicted bounding boxes are well aligned to ground truths. In particular, DiscoNet~\cite{li2021learning}, V2X-ViT~\cite{xu2022v2x}, and Where2comm~\cite{hu2022where2comm} fail to identify fast-moving objects (\ie, more ground truth bounding boxes are not matched).  In contrast, our method achieves holistic perception across heterogeneous agents by effectively aggregating the salient information in the spatio-temporal domains.

\noindent\textbf{Visualization of Attention Map.}
We visualize the multi-scale deformable cross-attention in the CCC component. As Figure~\ref{vis_attention} shows, our tailored module can dynamically aggregate the semantics of multiple sampled keypoints at different scales to enhance the visual representation of the reference point. Concretely, the deformable cross-attention learns to extract object-related valuable perception cues at lower scales by sampling keypoints inside or around targets. In comparison, it reasonably samples long-range features at higher scales to capture spatially critical information.

\vspace{-3pt}
\section{Conclusion}
In this paper, we present {\tt SCOPE}, a learning-based framework to address existing multi-agent collaborative perception challenges in an end-to-end manner. Our approach is the first time considering the temporal semantics from ego agents to capture valuable context cues through the context-aware component. Moreover, two spatial information aggregation components are introduced to achieve comprehensive information interaction among agents and refined feature fusion of ego agents. Extensive experiments on diverse 3D detection datasets prove the superiority of {\tt SCOPE} and the effectiveness of our components. 
The future work focuses on extending {\tt SCOPE} into more complex perception scenarios.

\section*{Acknowledgements}
This work is supported by the Shanghai Key Research Laboratory of NSAI, the National Natural Science Foundation of China (Grant No.\,62250410368), and the Nanjing First Automobile Works Grant.

{\small
\bibliographystyle{ieee_fullname}
\bibliography{iccv}

\begin{thebibliography}{10}\itemsep=-1pt

\bibitem{chen2019fcooper}
Qi Chen, Xu Ma, Sihai Tang, Jingda Guo, Qing Yang, and Song Fu.
\newblock F-cooper: Feature based cooperative perception for autonomous vehicle
  edge computing system using {{3D}} point clouds.
\newblock In {\em Proceedings of {{ACM}}/{{IEEE Symposium}} on {{Edge
  Computing}}}, pages 88--100, {Arlington Virginia}, Nov. 2019.

\bibitem{chen2019cooper}
Qi Chen, Sihai Tang, Qing Yang, and Song Fu.
\newblock Cooper: Cooperative perception for connected autonomous vehicles
  based on 3d point clouds.
\newblock In {\em 2019 IEEE 39th International Conference on Distributed
  Computing Systems (ICDCS)}, pages 514--524, 2019.

\bibitem{das2019tarmac}
Abhishek Das, Th{\'e}ophile Gervet, Joshua Romoff, Dhruv Batra, Devi Parikh,
  Mike Rabbat, and Joelle Pineau.
\newblock Tarmac: Targeted multi-agent communication.
\newblock In {\em International Conference on Machine Learning (ICML)}, pages
  1538--1546, 2019.

\bibitem{dosovitskiy2017carla}
Alexey Dosovitskiy, German Ros, Felipe Codevilla, Antonio Lopez, and Vladlen
  Koltun.
\newblock Carla: An open urban driving simulator.
\newblock In {\em Conference on Robot Learning (CoRL)}, pages 1--16, 2017.

\bibitem{du2021learning}
Yangtao Du, Dingkang Yang, Peng Zhai, Mingchen Li, and Lihua Zhang.
\newblock Learning associative representation for facial expression
  recognition.
\newblock In {\em Proceedings of the IEEE International Conference on Image
  Processing (ICIP)}, pages 889--893, 2021.

\bibitem{el2019rgb}
Khaled El~Madawi, Hazem Rashed, Ahmad El~Sallab, Omar Nasr, Hanan Kamel, and
  Senthil Yogamani.
\newblock Rgb and lidar fusion based 3d semantic segmentation for autonomous
  driving.
\newblock In {\em Proceedings of the IEEE Intelligent Transportation Systems
  Conference (ITSC)}, pages 7--12, 2019.

\bibitem{foerster2016learning}
Jakob Foerster, Ioannis~Alexandros Assael, Nando De~Freitas, and Shimon
  Whiteson.
\newblock Learning to communicate with deep multi-agent reinforcement learning.
\newblock {\em Advances in Neural Information Processing Systems (NIPS)}, 29,
  2016.

\bibitem{hochreiter1997long}
Sepp Hochreiter and J{\"u}rgen Schmidhuber.
\newblock Long short-term memory.
\newblock {\em Neural Computation}, 9(8):1735--1780, 1997.

\bibitem{hoshen2017vain}
Yedid Hoshen.
\newblock Vain: Attentional multi-agent predictive modeling.
\newblock {\em Advances in Neural Information Processing Systems (NIPS)}, 30,
  2017.

\bibitem{hu2022where2comm}
Yue Hu, Shaoheng Fang, Zixing Lei, Yiqi Zhong, and Siheng Chen.
\newblock Where2comm: Communication-efficient collaborative perception via
  spatial confidence maps.
\newblock In {\em Advances in Neural Information Processing Systems (NIPS)},
  2022.

\bibitem{hu2023collaboration}
Yue Hu, Yifan Lu, Runsheng Xu, Weidi Xie, Siheng Chen, and Yanfeng Wang.
\newblock Collaboration helps camera overtake lidar in 3d detection.
\newblock {\em arXiv preprint arXiv:2303.13560}, 2023.

\bibitem{jiang2018learning}
Jiechuan Jiang and Zongqing Lu.
\newblock Learning attentional communication for multi-agent cooperation.
\newblock {\em Advances in Neural Information Processing Systems (NIPS)}, 31,
  2018.

\bibitem{kingma2015adam}
Diederik~P. Kingma and Jimmy Ba.
\newblock Adam: A method for stochastic optimization.
\newblock In {\em International Conference on Learning Representations (ICLR)},
  2015.

\bibitem{lang2019pointpillars}
Alex~H Lang, Sourabh Vora, Holger Caesar, Lubing Zhou, Jiong Yang, and Oscar
  Beijbom.
\newblock Pointpillars: Fast encoders for object detection from point clouds.
\newblock In {\em Proceedings of the IEEE/CVF Conference on Computer Vision and
  Pattern Recognition (CVPR)}, pages 12697--12705, 2019.

\bibitem{lei2022latency}
Zixing Lei, Shunli Ren, Yue Hu, Wenjun Zhang, and Siheng Chen.
\newblock Latency-aware collaborative perception.
\newblock In {\em Proceedings of the European Conference on Computer Vision
  (ECCV)}, pages 316--332, 2022.

\bibitem{li2023learning}
Jinlong Li, Runsheng Xu, Xinyu Liu, Jin Ma, Zicheng Chi, Jiaqi Ma, and Hongkai
  Yu.
\newblock Learning for vehicle-to-vehicle cooperative perception under lossy
  communication.
\newblock {\em IEEE Transactions on Intelligent Vehicles}, 2023.

\bibitem{li2021learning}
Yiming Li, Shunli Ren, Pengxiang Wu, Siheng Chen, Chen Feng, and Wenjun Zhang.
\newblock Learning distilled collaboration graph for multi-agent perception.
\newblock {\em Advances in Neural Information Processing Systems (NIPS)},
  34:29541--29552, 2021.

\bibitem{liang2018deep}
Ming Liang, Bin Yang, Shenlong Wang, and Raquel Urtasun.
\newblock Deep continuous fusion for multi-sensor 3d object detection.
\newblock In {\em Proceedings of the European Conference on Computer Vision
  (ECCV)}, pages 641--656, 2018.

\bibitem{lin2017focal}
Tsung-Yi Lin, Priya Goyal, Ross Girshick, Kaiming He, and Piotr Doll{\'a}r.
\newblock Focal loss for dense object detection.
\newblock In {\em Proceedings of the IEEE/CVF International Conference on
  Computer Vision (ICCV)}, pages 2980--2988, 2017.

\bibitem{liu2020when2com}
Yen-Cheng Liu, Junjiao Tian, Nathaniel Glaser, and Zsolt Kira.
\newblock When2com: Multi-agent perception via communication graph grouping.
\newblock In {\em Proceedings of the IEEE/CVF Conference on Computer Vision and
  Pattern Recognition (CVPR)}, pages 4106--4115, 2020.

\bibitem{luo2022complementarity}
Guiyang Luo, Hui Zhang, Quan Yuan, and Jinglin Li.
\newblock Complementarity-enhanced and redundancy-minimized collaboration
  network for multi-agent perception.
\newblock In {\em Proceedings of the ACM International Conference on Multimedia
  (ACM MM)}, pages 3578--3586, 2022.

\bibitem{paszke2019pytorch}
Adam Paszke, Sam Gross, Francisco Massa, Adam Lerer, James Bradbury, Gregory
  Chanan, Trevor Killeen, Zeming Lin, Natalia Gimelshein, Luca Antiga, et~al.
\newblock Pytorch: An imperative style, high-performance deep learning library.
\newblock {\em Advances in Neural Information Processing Systems (NIPS)}, 32,
  2019.

\bibitem{ram2016effect}
Tika Ram and Khem Chand.
\newblock Effect of drivers’ risk perception and perception of driving tasks
  on road safety attitude.
\newblock {\em Transportation Research Part F: Traffic Psychology and
  Behaviour}, 42:162--176, 2016.

\bibitem{rauch2012car2x}
Andreas Rauch, Felix Klanner, Ralph Rasshofer, and Klaus Dietmayer.
\newblock Car2x-based perception in a high-level fusion architecture for
  cooperative perception systems.
\newblock In {\em IEEE Intelligent Vehicles Symposium}, pages 270--275, 2012.

\bibitem{singh2018learning}
Amanpreet Singh, Tushar Jain, and Sainbayar Sukhbaatar.
\newblock Learning when to communicate at scale in multiagent cooperative and
  competitive tasks.
\newblock {\em arXiv preprint arXiv:1812.09755}, 2018.

\bibitem{sukhbaatar2016learning}
Sainbayar Sukhbaatar, Rob Fergus, et~al.
\newblock Learning multiagent communication with backpropagation.
\newblock {\em Advances in Neural Information Processing Systems (NIPS)}, 29,
  2016.

\bibitem{tan1993multi}
Ming Tan.
\newblock Multi-agent reinforcement learning: Independent vs. cooperative
  agents.
\newblock In {\em International Conference on Machine Learning (ICML)}, pages
  330--337, 1993.

\bibitem{wang2020v2vnet}
Tsun-Hsuan Wang, Sivabalan Manivasagam, Ming Liang, Bin Yang, Wenyuan Zeng, and
  Raquel Urtasun.
\newblock V2vnet: Vehicle-to-vehicle communication for joint perception and
  prediction.
\newblock In {\em Proceedings of the European Conference on Computer Vision
  (ECCV)}, pages 605--621, 2020.

\bibitem{xiang2023v2xp}
Hao Xiang, Runsheng Xu, Xin Xia, Zhaoliang Zheng, Bolei Zhou, and Jiaqi Ma.
\newblock V2xp-asg: Generating adversarial scenes for vehicle-to-everything
  perception.
\newblock In {\em IEEE International Conference on Robotics and Automation
  (ICRA)}, pages 3584--3591. IEEE, 2023.

\bibitem{zhu2021deformable}
Zhu Xizhou, Su Weijie, Lu Lewei, Li Bin, Wang Xiaogang, and Dai Jifeng.
\newblock Deformable detr: Deformable transformers for end-to-end object
  detection.
\newblock In {\em International Conference on Learning Representations (ICLR)},
  2021.

\bibitem{xu2023model}
Runsheng Xu, Weizhe Chen, Hao Xiang, Xin Xia, Lantao Liu, and Jiaqi Ma.
\newblock Model-agnostic multi-agent perception framework.
\newblock In {\em 2023 IEEE International Conference on Robotics and Automation
  (ICRA)}, pages 1471--1478. IEEE, 2023.

\bibitem{xu2021opencda}
Runsheng Xu, Yi Guo, Xu Han, Xin Xia, Hao Xiang, and Jiaqi Ma.
\newblock Opencda: an open cooperative driving automation framework integrated
  with co-simulation.
\newblock In {\em 2021 IEEE International Intelligent Transportation Systems
  Conference (ITSC)}, pages 1155--1162. IEEE, 2021.

\bibitem{xu2022bridging}
Runsheng Xu, Jinlong Li, Xiaoyu Dong, Hongkai Yu, and Jiaqi Ma.
\newblock Bridging the domain gap for multi-agent perception.
\newblock {\em arXiv preprint arXiv:2210.08451}, 2022.

\bibitem{xu2022cobevt}
Runsheng Xu, Zhengzhong Tu, Hao Xiang, Wei Shao, Bolei Zhou, and Jiaqi Ma.
\newblock Cobevt: Cooperative bird's eye view semantic segmentation with sparse
  transformers.
\newblock In {\em Conference on Robot Learning (CoRL)}, 2022.

\bibitem{xu2023v2v4real}
Runsheng Xu, Xin Xia, Jinlong Li, Hanzhao Li, Shuo Zhang, Zhengzhong Tu,
  Zonglin Meng, Hao Xiang, Xiaoyu Dong, Rui Song, et~al.
\newblock V2v4real: A real-world large-scale dataset for vehicle-to-vehicle
  cooperative perception.
\newblock In {\em Proceedings of the IEEE/CVF Conference on Computer Vision and
  Pattern Recognition (CVPR)}, pages 13712--13722, 2023.

\bibitem{xu2022v2x}
Runsheng Xu, Hao Xiang, Zhengzhong Tu, Xin Xia, Ming-Hsuan Yang, and Jiaqi Ma.
\newblock V2x-vit: Vehicle-to-everything cooperative perception with vision
  transformer.
\newblock In {\em Proceedings of the European Conference on Computer Vision
  (ECCV)}, 2022.

\bibitem{xu2022opv2v}
Runsheng Xu, Hao Xiang, Xin Xia, Xu Han, Jinlong Li, and Jiaqi Ma.
\newblock Opv2v: An open benchmark dataset and fusion pipeline for perception
  with vehicle-to-vehicle communication.
\newblock In {\em Proceedings of the International Conference on Robotics and
  Automation (ICRA)}, pages 2583--2589, 2022.

\bibitem{yang2023context}
Dingkang Yang, Zhaoyu Chen, Yuzheng Wang, Shunli Wang, Mingcheng Li, Siao Liu,
  Xiao Zhao, Shuai Huang, Zhiyan Dong, Peng Zhai, and Lihua Zhang.
\newblock Context de-confounded emotion recognition.
\newblock In {\em Proceedings of the IEEE/CVF Conference on Computer Vision and
  Pattern Recognition (CVPR)}, pages 19005--19015, June 2023.

\bibitem{yang2022disentangled}
Dingkang Yang, Shuai Huang, Haopeng Kuang, Yangtao Du, and Lihua Zhang.
\newblock Disentangled representation learning for multimodal emotion
  recognition.
\newblock In {\em Proceedings of the 30th ACM International Conference on
  Multimedia (ACM MM)}, page 1642–1651, 2022.

\bibitem{yang2022contextual}
Dingkang Yang, Shuai Huang, Yang Liu, and Lihua Zhang.
\newblock Contextual and cross-modal interaction for multi-modal speech emotion
  recognition.
\newblock {\em IEEE Signal Processing Letters}, 29:2093--2097, 2022.

\bibitem{yang2022emotion}
Dingkang Yang, Shuai Huang, Shunli Wang, Yang Liu, Peng Zhai, Liuzhen Su,
  Mingcheng Li, and Lihua Zhang.
\newblock Emotion recognition for multiple context awareness.
\newblock In {\em Proceedings of the European Conference on Computer Vision
  (ECCV)}, volume 13697, pages 144--162. Springer, 2022.

\bibitem{yang2022learning}
Dingkang Yang, Haopeng Kuang, Shuai Huang, and Lihua Zhang.
\newblock Learning modality-specific and -agnostic representations for
  asynchronous multimodal language sequences.
\newblock In {\em Proceedings of the 30th ACM International Conference on
  Multimedia (ACM MM)}, page 1708–1717, 2022.

\bibitem{yang2023target}
Dingkang Yang, Yang Liu, Can Huang, Mingcheng Li, Xiao Zhao, Yuzheng Wang, Kun
  Yang, Yan Wang, Peng Zhai, and Lihua Zhang.
\newblock Target and source modality co-reinforcement for emotion understanding
  from asynchronous multimodal sequences.
\newblock {\em Knowledge-Based Systems}, page 110370, 2023.

\bibitem{yang2023novel}
Kun Yang, Jing Liu, Dingkang Yang, Hanqi Wang, Peng Sun, Yanni Zhang, Yan Liu,
  and Liang Song.
\newblock A novel efficient multi-view traffic-related object detection
  framework.
\newblock In {\em IEEE International Conference on Acoustics, Speech and Signal
  Processing (ICASSP)}, pages 1--5, 2023.

\bibitem{yang2022novel}
Kun Yang, Peng Sun, Jieyu Lin, Azzedine Boukerche, and Liang Song.
\newblock A novel distributed task scheduling framework for supporting
  vehicular edge intelligence.
\newblock In {\em IEEE International Conference on Distributed Computing
  Systems (ICDCS)}, pages 972--982, 2022.

\bibitem{yu2022dair}
Haibao Yu, Yizhen Luo, Mao Shu, Yiyi Huo, Zebang Yang, Yifeng Shi, Zhenglong
  Guo, Hanyu Li, Xing Hu, Jirui Yuan, et~al.
\newblock Dair-v2x: A large-scale dataset for vehicle-infrastructure
  cooperative 3d object detection.
\newblock In {\em Proceedings of the IEEE/CVF Conference on Computer Vision and
  Pattern Recognition (CVPR)}, pages 21361--21370, 2022.

\bibitem{yuan2021temporal}
Zhenxun Yuan, Xiao Song, Lei Bai, Zhe Wang, and Wanli Ouyang.
\newblock Temporal-channel transformer for 3d lidar-based video object
  detection for autonomous driving.
\newblock {\em IEEE Transactions on Circuits and Systems for Video Technology},
  32(4):2068--2078, 2021.

\bibitem{zhang2021safe}
Zixu Zhang and Jaime~F Fisac.
\newblock Safe occlusion-aware autonomous driving via game-theoretic active
  perception.
\newblock {\em arXiv preprint arXiv:2105.08169}, 2021.

\bibitem{zhou2022centerformer}
Zixiang Zhou, Xiangchen Zhao, Yu Wang, Panqu Wang, and Hassan Foroosh.
\newblock Centerformer: Center-based transformer for 3d object detection.
\newblock In {\em Proceedings of the European Conference on Computer Vision
  (ECCV)}, pages 496--513, 2022.

\end{thebibliography}
}

\end{document}